\documentclass[11pt]{article}

\usepackage[margin=1in]{geometry}
\usepackage[T1]{fontenc}
\usepackage[utf8]{inputenc}
\usepackage{lmodern}
\usepackage{microtype}
\usepackage{amsmath,amssymb}
\usepackage{booktabs}
\usepackage{tabularx}
\usepackage{array}
\usepackage{xcolor}
\usepackage{graphicx}
\usepackage{tikz}
\usetikzlibrary{arrows.meta,positioning,fit,calc}
\usepackage[numbers,sort&compress]{natbib}
\usepackage{xurl}
\usepackage[colorlinks=true,allcolors=blue!55!black]{hyperref}

\definecolor{accent}{HTML}{164A6E}
\definecolor{softblue}{HTML}{EAF3F8}
\definecolor{softgreen}{HTML}{EAF5EF}
\definecolor{softorange}{HTML}{FFF2DF}
\definecolor{softred}{HTML}{FBE9E7}
\newcolumntype{Y}{>{\raggedright\arraybackslash}X}
\newcolumntype{L}[1]{>{\raggedright\arraybackslash}p{#1}}
\title{From Language Models to World-Acting Systems: Progress and Limits of Agentic AI across Digital, Social, Virtual, and Physical Environments}
\author{Linsen Zhu \and Mengqing Cai}
\date{Literature cutoff: 31 August 2026}

\begin{document}
\hypersetup{
  pdftitle={From Language Models to World-Acting Systems: Progress and Limits of Agentic AI across Digital, Social, Virtual, and Physical Environments},
  pdfauthor={Linsen Zhu and Mengqing Cai},
  pdfsubject={A critical state-of-the-art review of agentic AI across digital, social, virtual, and physical environments},
  pdfkeywords={agentic AI, language-model agents, tool use, computer use, multi-agent systems, world models, robotics, human oversight, AI safety}
}
\maketitle

\begin{abstract}
Large language models become consequential agents when surrounding systems let outputs change external state. Models now call tools, operate interfaces, delegate work, retain state, inhabit generated worlds, and control robots or laboratory equipment. Such advances are often narrated as one march toward autonomy, conflating model competence, system integration, persistence, and safe authority. This critical review synthesizes primary research and official technical specifications available by 31 August 2026. We organize the evidence along delegated authority, temporal persistence, and environmental coupling, while separating model, harness, and environment. Within the evidence examined, action-interface expansion is documented more convincingly than robust completion, recovery, authorization, or independent verification. Model Context Protocol and Agent2Agent improve interoperability but do not establish trustworthy delegation; multi-agent organization adds specialization alongside cost and correlated failure. Persistent simulations and world models support training and planning but do not themselves demonstrate agency; robotics and self-driving laboratories establish bounded feasibility rather than unattended open-world reliability. We propose \emph{justified delegation} as an analytical and normative heuristic, not an observed law or certified score: expand action scope only where evidence supports provenance, bounded authority, failure detection, safe recovery, and calibrated human control. This framing yields a research agenda for coupled model--harness evaluation, capability-based permissions, durable state, cross-agent accountability, and staged physical validation.

\end{abstract}


\section{Introduction}

Language models were first encountered chiefly as generators: a user supplied text and received text. Agentic systems alter the practical meaning of that exchange. A model output may now trigger a database query, edit a code repository, navigate a website, schedule another agent, ask a person to resolve an ambiguity, or move a physical instrument. The central scientific question is therefore no longer whether a model can produce a plausible plan. It is whether a configured system can execute delegated work over time while preserving the user's intent, institutional constraints, and a recoverable account of what happened.

Early agent research established several ingredients of this transition. ReAct interleaved reasoning traces with actions and environmental observations, Toolformer studied learned decisions about when and how to call application programming interfaces (APIs), and Reflexion used linguistic feedback to revise subsequent behaviour \citep{yao2023react,schick2023toolformer,shinn2023reflexion}. Cognitive-architecture accounts subsequently made explicit that memory, action selection, and decision procedures sit around rather than wholly inside a foundation model \citep{sumers2024coala}. These contributions changed the unit of analysis. An agent is not simply an especially capable model; it is a model configured within a control loop, an interface, a state representation, and a set of permissions.

Since 2023, each part of this surrounding system has broadened. Web and desktop benchmarks introduced executable environments with functional end-state evaluation \citep{zhou2024webarena,xie2024osworld}. Coding agents showed that an agent-computer interface can materially affect performance even when the underlying model is held fixed \citep{yang2024sweagent}. Multi-agent frameworks made delegation and role specialization programmable \citep{wu2024autogen}. Open protocols now expose tools and resources to model applications and define task exchange between otherwise opaque agents \citep{anthropic2024mcp,a2a2026spec}. Persistent simulations, video world models, and generated interactive environments seek to maintain or predict world state \citep{park2023generativeagents,sima2024,assran2025vjepa2,vast2026eden}. Vision-language-action models and laboratory systems connect high-level instructions to robot or instrument actions \citep{zitkovich2023rt2,kim2024openvla,boiko2023coscientist,bran2024chemcrow}. In August 2026, Anthropic's Model Hardware Standard preview made the architectural convergence unusually visible by proposing a model-agnostic driver layer through which agents could discover and operate heterogeneous programmable devices \citep{anthropic2026mhs}.

This breadth creates a conceptual problem. ``More agentic'' is frequently used to describe at least four different changes: a stronger base model, a richer software harness, a longer-lived task, or access to a more consequential environment. These changes do not imply one another. A model can reason well but possess no authority to act. A weak model can be given broad credentials. A protocol can make tools interoperable without making their use safe. A visually coherent world model can supply an environment while containing no goal-directed agent. A robot policy can produce competent motor actions without maintaining a long-term objective or negotiating authority. Treating these systems as points on a single autonomy ladder therefore hides the variables that most strongly determine both usefulness and risk.

Existing surveys have usefully organized autonomous language-model agents by their components and applications, or concentrated specifically on how agents are evaluated \citep{wang2024survey,yehudai2026evaluation}. This review does not claim a new inventory of agent components. Its distinct synthesis compares otherwise separated domains through delegated authority, temporal persistence, and environmental coupling, while keeping model, harness, environment, delegator, and evidence source analytically separate. That combination is intended to expose assurance gaps that an architecture- or benchmark-centred account can leave implicit.

The empirical record also resists a simple progress narrative. In the original WebArena study, the strongest reported GPT-4 agent completed 14.41\% of tasks, compared with 78.24\% for humans; the original OSWorld study reported performance below 12.2\% for its best configuration and 72.4\% for humans \citep{zhou2024webarena,xie2024osworld}. Later first-party computer-use systems reported substantial gains, but still acknowledged reliability limitations and the need for human oversight \citep{openai2025cua,openai2025operatorcard}. In tool-mediated service tasks, $\tau$-bench showed that a system that sometimes completes a trajectory can be much less reliable when success is required consistently across repeated trials \citep{yao2025taubench}. Multi-agent debate can drift away from the original problem over longer interactions \citep{becker2026drift}, and equal-compute comparisons question whether some reported multi-agent gains arise from organization rather than additional inference budget \citep{tran2026singleagent}. Physical systems add sensing uncertainty, actuation error, wear, safety envelopes, and consequences that cannot always be reset.

This review advances one evidence-bounded inference: \emph{within the public evidence examined here, action interfaces have expanded more convincingly than evidence for verifiable autonomy}. We use ``autonomy'' narrowly, to mean sustained goal-directed control in which a system selects and revises actions under delegated authority. We use ``verifiable'' to mean that success, policy compliance, and important side effects can be checked by evidence outside the model's own narrative. This comparison is not a measured field-wide growth law, nor does it imply that models have made little progress. It instead records an asymmetry across the representative benchmarks, experiments, specifications, and first-party artifacts reviewed here: capability and interface gains are frequently visible, whereas verification, recovery, and governance remain external, incomplete, or evaluated only in resettable settings.

To make this asymmetry legible, we organize agentic systems along three independent dimensions. \textbf{Delegated authority} asks what state changes the system is permitted to initiate and whose authorization is required. \textbf{Temporal persistence} asks whether objectives, memory, credentials, and obligations survive across steps, interruptions, or episodes. \textbf{Environmental coupling} asks how directly actions affect software replicas, live digital services, shared social processes, simulated worlds, or physical systems. These dimensions complement an explicit separation between the \emph{model}, the \emph{harness} that manages context and action, and the \emph{environment} that supplies state and consequences.

The resulting synthesis contributes three things. First, it connects research that is usually reviewed separately---tool use, computer-use agents, multi-agent systems, human--agent interaction, world models, robotics, and autonomous laboratories---without claiming that they form a single technical lineage. Second, it grades conclusions by source type and evaluation setting, separating peer-reviewed benchmark or experimental evidence from preprints, open protocol specifications, and company research previews. Third, it derives a research agenda centred on justified delegation rather than maximal independence. This is a critical review, not a systematic review or a quantitative meta-analysis: it aims to clarify concepts, compare evidential strength, and identify unresolved dependencies rather than estimate corpus-wide prevalence.

\section{Scope, review method, and evidential discipline}

\subsection{A critical rather than systematic review}

The agent literature is heterogeneous in both object and evidence. It includes model-training papers, systems papers, benchmarks, protocol specifications, open-source frameworks, product documentation, system cards, and demonstrations whose primary output is a video or blog post. A narrowly bibliometric review would underrepresent standards and deployed interfaces, whereas treating every public artifact as equivalent would turn announcements into scientific results. We therefore use a critical, question-driven review design. The organizing question is: \emph{what evidence supports the transfer of delegated control from language models into digital, social, virtual, and physical environments, and where does that evidence stop?}

The literature cutoff is 31 August 2026. We prioritize original peer-reviewed papers and official proceedings records for capability and benchmark claims; public preprints or institutional technical reports when no archival version existed by the cutoff; the maintained specification and release record for protocol semantics; and first-party research announcements or system cards only for claims about previewed systems, interfaces, and disclosed limitations. Surveys are used to position terminology, not as substitutes for a primary result \citep{wang2024survey,yehudai2026evaluation}. We include representative systems when they expose a transition in the action stack: reasoning--action loops, learned tool use, executable web or desktop interaction, inter-agent delegation, persistent state, interactive three-dimensional environments, robot control, or closed-loop laboratory operation. We exclude ordinary chat systems with no external action path and secondary reports that add no independently inspectable evidence.

This design does not support statements about the proportion of all agent papers with a given feature, nor does it yield a formal meta-analytic effect size. Model versions, prompts, inference budgets, scaffolds, and benchmark states change too quickly for many headline scores to be pooled. Instead, we use representative contrasts to test conceptual claims. Historical performance numbers are identified as results of a named configuration, not presented as a current leaderboard. Product availability is evidence that an interface was deployed, not that it was reliable. A specification is evidence of a defined message or authorization surface, not evidence that implementations comply or that participating agents are competent.

\subsection{Evidence categories and claim boundaries}

Table~\ref{tab:evidence} defines the source categories used throughout this review. They are not a single prestige ranking. A protocol specification is the strongest source for wire semantics but a poor source for task performance; a controlled physical experiment can establish feasibility without establishing longitudinal safety; and an independent benchmark can measure task success while omitting real credentials and affected users. We therefore attach evidence quality to a particular claim rather than assigning a permanent score to a system.

\begin{table*}[t]
\centering
\caption{Evidence categories used in this review and the claims each can support.}
\label{tab:evidence}
\small
\begin{tabularx}{\textwidth}{L{2.6cm}Y Y Y}
\toprule
\textbf{Source category} & \textbf{Strongest defensible use} & \textbf{Typical blind spot} & \textbf{Representative examples} \\
\midrule
Peer-reviewed benchmark or systems paper & Performance and failure under the published task, model, harness, budget, and evaluator & Distribution shift, live permissions, changing interfaces, and longitudinal incidents & WebArena, OSWorld, SWE-agent, $\tau$-bench \citep{zhou2024webarena,xie2024osworld,yang2024sweagent,yao2025taubench} \\
Peer-reviewed controlled physical experiment & Feasibility and measured behaviour in a disclosed apparatus and procedure & Scale, repeated unattended operation, rare hazards, and transfer to other instruments & RT-2, Coscientist, ChemCrow \citep{zitkovich2023rt2,boiko2023coscientist,bran2024chemcrow} \\
Preprint or institutional technical report & Timely architecture, released artifact, and provisional evaluation & Peer review, independent replication, stable model endpoints, and selective reporting & Voyager, V-JEPA~2, Gemini Robotics, Magentic-UI \citep{wang2023voyager,assran2025vjepa2,gemini2025robotics,mozannar2025magenticui} \\
Open technical specification & Defined roles, messages, state transitions, discovery, and security requirements & Implementation quality, adoption, semantic correctness, and end-task reliability & MCP and A2A \citep{mcp2026spec,a2a2026spec} \\
First-party preview, system card, or product record & Existence, stated design, first-party tests, deployment constraints, and disclosed failure modes & Independent replication, complete denominators, comparison fairness, and durability & OpenAI CUA, Genie~3, Project Eden, MHS \citep{openai2025cua,deepmind2025genie3,vast2026eden,anthropic2026mhs} \\
\bottomrule
\end{tabularx}
\parbox{\textwidth}{\footnotesize\emph{Note:} A source may move category when an archival paper, public specification, or independent evaluation appears. All category assignments describe status at the literature cutoff.}
\end{table*}

Three boundary rules prevent common inferential jumps. First, a demonstration of \emph{capability} does not establish a frequency of reliable success. A single completed trajectory can be valuable feasibility evidence while saying little about tail risk. Second, \emph{persistence} is decomposed into different objects. Conversation context, episodic memory, a resumable task record, a learned skill, and durable authority are not interchangeable. Third, \emph{openness} refers to the action environment, not visual richness. A generated landscape may be unbounded in appearance while running inside a resettable simulator; a short API call may affect a live bank, hospital, or public service.

\subsection{The unit of analysis: a configured action system}

For any reported result, the relevant unit is the configured system
\begin{equation}
\mathcal{S} = (M,H,E,U),
\end{equation}
where $M$ is the model, $H$ the harness, $E$ the operational environment, and $U$ the user or institution that delegates authority. The harness includes prompting, context construction, memory, tool schemas, planners, verifiers, retry logic, identity and credential handling, and user-interface controls. The environment supplies observable state, admissible actions, transition dynamics, and consequences. The delegator supplies the objective and legitimate authority, but may remain available for clarification, approval, or takeover. Affected people and institutions are not collapsed into $U$: they may experience consequences without delegating authority or operating the system, and therefore require separate channels for notice, contestation, and remedy. This formulation makes clear why ``model performance'' is often a system property: changing the accessible commands or feedback representation can alter outcomes without changing weights \citep{yang2024sweagent}.

We analyze $\mathcal{S}$ on three dimensions. Delegated authority ranges from recommendation, through reversible action under approval, to direct state-changing action. Temporal persistence ranges from a single response, through multi-step and resumable tasks, to recurring or cross-episode obligations. Environmental coupling ranges from synthetic or resettable state, through bounded live services, to shared social and physical environments with difficult-to-reverse effects. These dimensions are descriptive rather than celebratory. A high-coupling system is not necessarily more intelligent; it simply requires stronger evidence and controls for the same confidence in deployment.

\section{From model capability to configured agency}

\subsection{Reasoning and tool use are necessary but not sufficient}

The most important conceptual move in agentic AI is from predicting an answer to selecting an intervention. ReAct made this move explicit by alternating reasoning and action tokens with environmental observations, allowing a language model to gather information and revise its trajectory \citep{yao2023react}. Toolformer addressed a complementary problem: learning which API calls are useful and how their returned values should enter subsequent prediction \citep{schick2023toolformer}. Reflexion showed that textual feedback from failure can be retained and used to improve later attempts without weight updates \citep{shinn2023reflexion}. Together, these systems established a reusable loop: infer a next step, invoke an affordance, observe the result, and condition the next step on that observation.

The loop is easy to draw and hard to validate. A model must choose the right subgoal, bind arguments to an action schema, recognize when the observation is incomplete or adversarial, preserve relevant state, and stop for the right reason. More capable reasoning can improve several of these operations, but it does not determine the permissions of the invoked tool, the fidelity of the observation, or the correctness of an external side effect. Tool-use accuracy measured against a reference call also does not show that the action was authorized or that the desired final state was achieved.

Longer effective task horizons provide one indicator of improved competence. Kwa and colleagues define a 50\%-task-completion time horizon as the duration humans require for tasks that an AI system can complete with 50\% success, and report a rapid historical increase on their suite of software and research tasks \citep{kwa2025longtasks}. The metric is valuable because it converts a collection of benchmark results into an operationally interpretable scale and attributes gains partly to reliability, adaptation to mistakes, reasoning, and tool use. Its boundary is equally important: the tasks are comparatively clean, the 50\% threshold is below the reliability demanded in many deployments, and human completion time is a task-complexity proxy rather than the time an agent can safely remain unattended. A longer benchmark horizon therefore supports a capability claim, not a general claim of durable autonomy.

\subsection{The harness is part of the causal system}

The harness turns a model into an action system. At minimum, it defines the system prompt, available actions, observation encoding, context-window policy, and termination rule. Operational harnesses add memory stores, plan representations, sandboxing, credential brokers, rate and cost limits, approval gates, monitoring, retries, checkpoints, and recovery procedures. Cognitive Architectures for Language Agents organizes related components as memory, action space, and decision procedures, helping separate a language model's learned knowledge from the control architecture in which it operates \citep{sumers2024coala}. More recent harness-engineering work treats context construction and runtime policy as objects of optimization in their own right, although much of this literature remained preprint evidence at the cutoff \citep{ning2026codeharness,lin2026agenticharness}.

SWE-agent gives direct empirical support to the importance of interface design. Its agent-computer interface supplies model-oriented commands and feedback for repository navigation, editing, and testing; the original NeurIPS study reported a 12.5\% pass@1 rate on SWE-bench and argued that interface choices materially shape behaviour \citep{yang2024sweagent}. The number is now historical, but the inference remains: a benchmark score cannot be assigned to model weights alone when a redesigned interaction surface changes what the same class of model can accomplish.

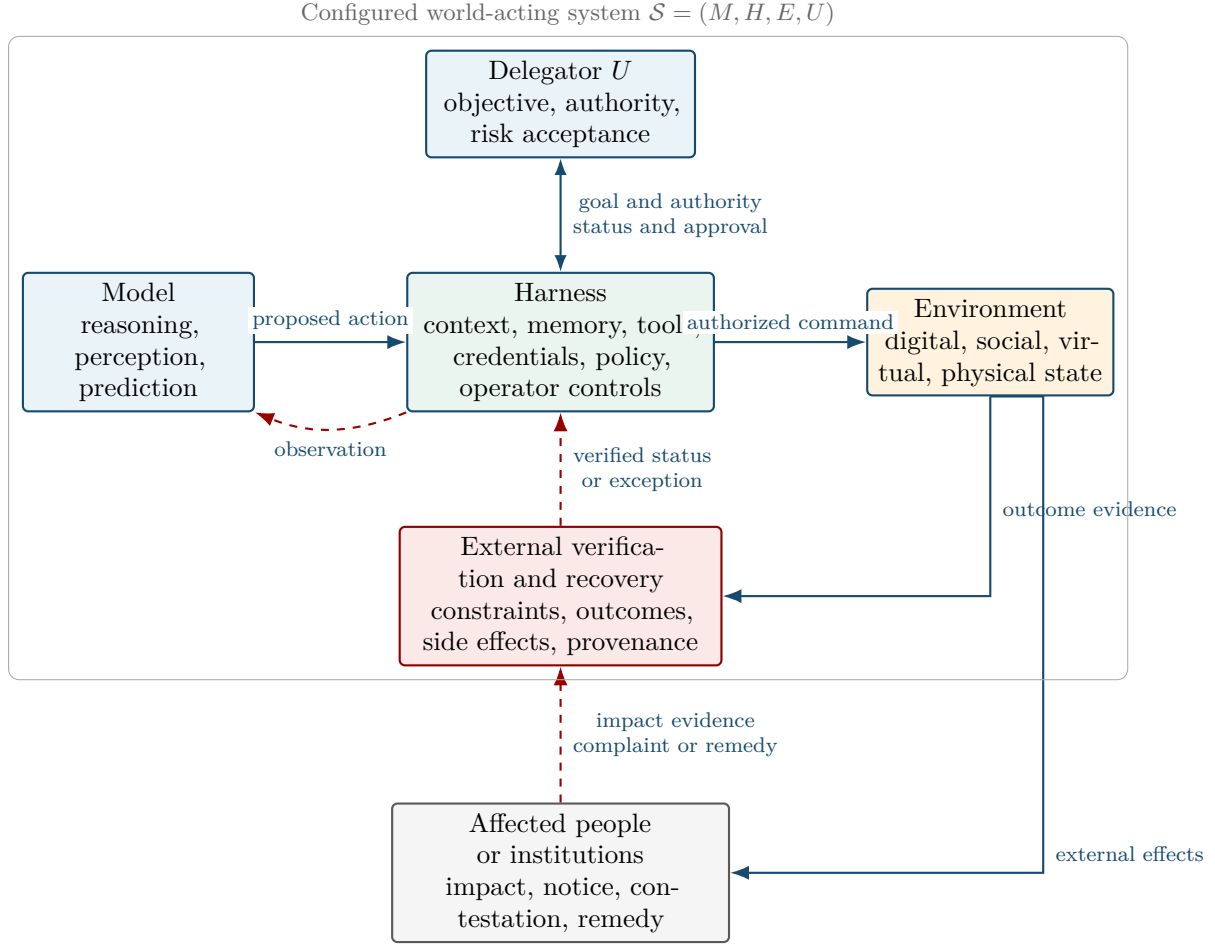
\begin{figure*}[t]
\centering
\begin{tikzpicture}[
  node distance=15mm and 20mm,
  box/.style={draw=accent, rounded corners=2pt, thick, minimum height=11mm, align=center, fill=softblue},
  hbox/.style={draw=accent, rounded corners=2pt, thick, minimum height=11mm, align=center, fill=softgreen},
  ebox/.style={draw=accent, rounded corners=2pt, thick, minimum height=11mm, align=center, fill=softorange},
  vbox/.style={draw=red!55!black, rounded corners=2pt, thick, minimum height=10mm, align=center, fill=softred},
  sbox/.style={draw=gray!70!black, rounded corners=2pt, thick, minimum height=10mm, align=center, fill=gray!8},
  arr/.style={-{Latex[length=2.5mm]}, thick, draw=accent},
  back/.style={-{Latex[length=2.5mm]}, thick, dashed, draw=red!60!black},
  exchange/.style={{Latex[length=2.2mm]}-{Latex[length=2.2mm]}, thick, draw=accent},
  edge/.style={fill=white, inner sep=1.5pt, font=\scriptsize, align=center, text=accent},
  font=\small
]
\node[box, text width=2.8cm] (model) {Model\\reasoning, perception, prediction};
\node[hbox, right=of model, text width=3.8cm] (harness) {Harness\\context, memory, tools, credentials, policy, operator controls};
\node[ebox, right=of harness, text width=3.0cm] (env) {Environment\\digital, social, virtual, physical state};
\node[box, above=of harness, text width=3.3cm] (delegator) {Delegator $U$\\objective, authority, risk acceptance};
\node[vbox, below=of harness, text width=4.0cm] (verify) {External verification and recovery\\constraints, outcomes, side effects, provenance};
\node[sbox, below=18mm of verify, text width=4.2cm] (affected) {Affected people or institutions\\impact, notice, contestation, remedy};

\draw[arr] (model) -- node[edge, above=1mm]{proposed action} (harness);
\draw[back] (harness.south west) to[bend left=22] node[edge, below=1mm]{observation} (model.south east);
\draw[arr] (harness) -- node[edge, above=1mm]{authorized command} (env);
\draw[arr] (env.south) |- node[edge, pos=.28, right=1mm]{outcome evidence} (verify.east);
\draw[back] (verify.north) -- node[edge, right=1mm]{verified status\\or exception} (harness.south);
\draw[exchange] (delegator.south) -- node[edge, right=1mm]{goal and authority\\status and approval} (harness.north);
\draw[arr] (env.south) -- ++(7mm,0) |- node[edge, pos=.48, right=1mm]{external effects} (affected.east);
\draw[back] (affected.north) -- node[edge, right=1mm]{impact evidence\\complaint or remedy} (verify.south);

\node[draw=gray!65, rounded corners, inner sep=5pt, fit=(delegator)(model)(harness)(env)(verify), label={[font=\footnotesize,text=gray!80!black]above:Configured world-acting system $\mathcal{S}=(M,H,E,U)$}] {};
\end{tikzpicture}
\caption{Agency is a property of a configured model--harness--environment system under authority granted by the delegator $U$. Verification and recovery are drawn as an expanded function of the harness $H$, not as a fifth element of $\mathcal{S}$. An authorized human operator interacts through $H$ but is not equated with $U$; affected people and institutions need not have delegated or operated the system and are therefore shown outside its boundary. Their impacts, complaints, and remedies remain inputs to independent verification and recovery.}
\label{fig:system}
\end{figure*}

Figure~\ref{fig:system} summarizes this configured-system view. The delegator $U$ is the principal that grants an objective and legitimate authority; a human operator may carry out approval, monitoring, or takeover on that principal's behalf; and affected people or institutions may occupy neither role. The model proposes, the harness mediates, and the environment changes. Verification must compare observed state with the delegated objective and applicable constraints, rather than merely ask the same model whether it believes it succeeded. Keeping these actors separate avoids two symmetrical errors: treating every human intervention as evidence that the system is not agentic, or treating an available takeover button as proof of effective control.

\subsection{Three separable meanings of ``better''}

This decomposition yields three distinct improvement claims. \emph{Policy competence} improves when the model selects more appropriate actions or recovers from a wider class of errors under a fixed interface. \emph{Action coverage} improves when the harness exposes additional tools, applications, agents, or devices. \emph{Assurance} improves when authorization, verification, containment, and recovery become more reliable. Across the systems examined here, the first two are often documented more directly than the third. A model may gain a universal mouse-and-keyboard interface, for example, even though success remains difficult to verify and an indirect prompt injection can redirect the same interface.

Calling all three improvements ``autonomy'' creates misleading comparisons. A narrow coding agent with tests and a version-control diff may be more verifiable than a stronger general model navigating an account through pixels. A laboratory orchestrator that operates only allowlisted procedures behind interlocks may be more trustworthy than an apparently less consequential agent with unrestricted network and credential access. The appropriate research objective is therefore not to maximize action coverage in isolation, but to expand it at a rate supported by competence and assurance.

\section{Digital action: from typed APIs to universal computer use}

\subsection{Structured tools provide narrow, inspectable affordances}

APIs are the most legible action interface available to an agent. A typed call names an operation and its arguments; the service can authenticate the caller, validate a schema, record the request, and return structured state. This makes precondition checks and policy enforcement more tractable than when the same operation is reconstructed through a graphical interface. Learned and prompted tool-use systems demonstrated that language models can select among such functions, fill arguments, and incorporate returned observations \citep{schick2023toolformer,yao2023react}. AgentBench broadened evaluation across operating systems, databases, knowledge graphs, games, and embodied settings, documenting persistent weaknesses in long-horizon reasoning, decision making, and instruction following \citep{liu2024agentbench}.

Structured access does not by itself make an action correct. Tool descriptions can be ambiguous, arguments can be valid but semantically wrong, and a technically successful call can violate a user's unstated preference or an institutional policy. A read operation may disclose sensitive context to the model; a write operation may be non-idempotent; and a sequence of individually permitted calls may create an impermissible aggregate outcome. These are composition problems, not parsing problems. They motivate separation among the model's proposed action, the harness's policy decision, and the service's final authorization.

Transactional benchmarks make this distinction measurable. $\tau$-bench evaluates agents interacting with both users and domain tools in retail and airline scenarios, checking whether the final database state satisfies task-specific policies \citep{yao2025taubench}. Its pass$^k$ analysis showed that consistency across repeated attempts falls well below one-shot success. This is a central result for deployment: if an agent succeeds with probability $p$ on one run and errors are independent, requiring $k$ consecutive successes yields $p^k$; correlated failures or changing external state can make reliability worse. A persuasive transcript is therefore weaker evidence than a verified final state across repeated trials.

\subsection{Graphical interfaces expand reach by weakening structure}

Many consequential applications expose incomplete or inaccessible APIs. Computer-use agents instead perceive screenshots or accessibility trees and act with mouse and keyboard commands. This interface is attractive because it reaches legacy and proprietary software without a bespoke integration. It is also intrinsically less typed: visual grounding, focus state, timing, window layout, transient notifications, and hidden application state all affect the meaning of a click.

WebArena was an early step from static language evaluation to functional action evaluation. It provides reproducible, self-hosted websites in e-commerce, social discussion, collaborative development, and content management, with evaluators that inspect task outcomes rather than merely compare action strings. Its best reported GPT-4 baseline achieved 14.41\% end-to-end success against 78.24\% for humans \citep{zhou2024webarena}. OSWorld extended this approach to real applications in Ubuntu, Windows, and macOS through a resettable virtual-machine environment. The original study contained 369 tasks and reported less than 12.2\% success for the best evaluated agent compared with 72.4\% for humans \citep{xie2024osworld}. These results are historical baselines rather than current leaderboard estimates, but they remain strong evidence that perception, operational knowledge, and multi-application state tracking are distinct from fluent language generation.

First-party systems subsequently reported substantial benchmark gains. OpenAI's Computer-Using Agent (CUA) research preview reported 38.1\% on OSWorld and 58.1\% on WebArena with a screen--mouse--keyboard action space \citep{openai2025cua}. The associated system card explicitly described 38.1\% as insufficiently reliable for operating-system automation and recommended human oversight, while also identifying prompt injection, model mistakes, and jailbreak-related concerns \citep{openai2025operatorcard}. Anthropic similarly released computer use as a public beta and warned that the feature remained experimental and could make errors when interpreting screens or taking actions \citep{anthropic2024computeruse}. Such reports establish that general graphical control became an accessible interface; because the evaluation and deployment reports are first-party, they do not replace independent tests under stable harnesses.

The difference between APIs and graphical control is therefore not a choice between ``limited'' and ``general'' agency. It is a trade between explicit semantics and interface coverage. An API can expose high-impact operations with precise arguments; a graphical interface can reach almost any visible control while making state and intent harder to validate. Hybrid systems should prefer structured calls when their semantics and authorization are adequate, fall back to computer use when necessary, and retain the same policy and provenance layer across both. Otherwise the fallback becomes a path around controls rather than a compatibility mechanism.

\subsection{Coding agents illustrate both leverage and verifiability}

Software engineering is a particularly informative digital domain because actions can be consequential yet unusually inspectable. An agent can search a repository, edit files, execute tests, and propose a patch inside an isolated environment. Version control exposes a diff; tests offer partial executable specifications; and failed changes can often be reverted. SWE-bench created issue-level tasks from real GitHub repositories, while SWE-agent showed that model-oriented navigation and editing commands improve the interaction between a language model and a repository \citep{jimenez2024swebench,yang2024sweagent}.

This domain has also produced deployed asynchronous agents that accept tasks, work in isolated environments, and return changes for review. OpenAI's Codex launch and subsequent general-availability documentation describe cloud task delegation, parallel work, resumable sessions, software-development-kit access, and administrative controls \citep{openai2025codex,openai2025codexga}. These are product records rather than controlled comparative evidence. Their scientific importance lies in the system pattern: the unit of delegation is a durable task with an auditable artifact, not a stream of chat turns.

Even here, verifiability is incomplete. Test suites encode only part of intended behaviour; generated tests can share the agent's mistaken assumptions; dependency installation and network access enlarge the attack surface; and a plausible diff can introduce security or maintenance debt outside the tested path. Coding agents are thus not an exception to the action--verification gap. They show how environmental affordances---sandboxing, version control, test execution, and human review---can narrow it.

\subsection{Digital action is already social action}

The digital and social categories overlap whenever software mediates people, money, reputation, or access. Sending a message, cancelling a booking, modifying a shared document, or posting in a community changes other people's environment. Benchmark replicas correctly isolate these effects for reproducibility, but that isolation removes affected parties, evolving terms of service, account recovery, and institutional appeals. A claim about success in a self-hosted web replica should therefore not be generalized to safe operation on arbitrary live accounts. The relevant transition is not from ``web'' to ``real world''; it is from resettable state with enumerable stakeholders to live state whose consequences propagate beyond the benchmark.

\section{Delegation, persistence, and multi-agent organization}

\subsection{Protocols standardize exchange, not intelligence}

As agents gained tools and longer tasks, integration shifted from embedding every capability in one application to connecting independently developed components. Two protocols became prominent but solve different problems. The Model Context Protocol (MCP), introduced by Anthropic in 2024 and subsequently developed as an open specification, connects an AI application to servers that expose tools, resources, and prompts \citep{anthropic2024mcp,mcp2026spec}. Agent2Agent (A2A), announced by Google in 2025 and governed as a Linux Foundation project, supports discovery and task exchange between opaque agents \citep{google2025a2a,linuxfoundation2025a2a,a2a2026spec}. MCP is principally a model-application-to-capability interface; A2A is principally an agent-to-agent work protocol.

The distinction matters because ``agents can communicate'' can describe very different guarantees. MCP defines how a host learns what a server offers and invokes it. The 28 July 2026 revision adopted a stateless core, explicit discovery, cacheable list results, multi-round-trip requests, authorization hardening, and an extension mechanism; long-running tasks moved to an extension \citep{mcp2026spec,mcp2026release}. Stateless transport does not require stateless applications: a tool can return an explicit handle that the model carries across calls. A2A version 1.0 defines Agent Cards for discovery, messages and artifacts for exchange, and stateful tasks with lifecycle states, streaming, and push-notification options \citep{a2a2026spec}. A task can enter an \texttt{input-required} state and continue under the same identifiers after a client provides additional information.

Neither protocol establishes that a discovered capability is trustworthy, that an agent understands another agent's intent, or that a completed status corresponds to the delegator's desired world state. Authentication can establish an identity or credential relationship; it does not validate the semantics of a proposed task. Signed capability metadata can protect integrity; it does not demonstrate competence. Protocol adoption is thus evidence of a maturing action substrate, not evidence that autonomous collaboration is solved.

\begin{table*}[t]
\centering
\caption{Different delegation interfaces expose complementary semantics and leave complementary gaps.}
\label{tab:delegation}
\small
\begin{tabularx}{\textwidth}{L{2.5cm}Y Y Y Y}
\toprule
\textbf{Interface} & \textbf{Primary relationship} & \textbf{State / continuation} & \textbf{Human-control affordance} & \textbf{What it does not establish} \\
\midrule
MCP 2026-07-28 & AI host or client to tool/resource server & Stateless protocol core; explicit handles; task extension for long work & Multi-round-trip input or confirmation mediated by the client & Tool safety, least privilege, semantic success, or server trustworthiness \citep{mcp2026spec,mcp2026release} \\
A2A 1.0 & Client agent to remote, internally opaque agent & Stateful task lifecycle, context identifiers, artifacts, streaming and push & \texttt{input-required}, cancellation, authenticated task continuation & Benefit of multiple agents, truthful capability claims, or correctness of completed artifacts \citep{a2a2026spec} \\
Human-centred agent UI & Human delegator to one or more software agents & Plans, parallel sessions, checkpoints, and optional memory & Co-planning, co-tasking, action approval, intervention, and final verification & That people notice every hazard, understand approvals, or can supervise at scale \citep{mozannar2025magenticui} \\
Physical driver layer & Agent or program to heterogeneous programmable devices & Device state, commands, streamed measurements, and reusable scripts & Hardware limits, allowlists, supervision, and escalation depend on implementation & Physical reasoning, calibrated sensing, general fault recovery, or unattended safety \citep{anthropic2026mhs} \\
\bottomrule
\end{tabularx}
\end{table*}

\subsection{Agent-to-human delegation is an escalation pattern, not a settled protocol}

Delegation is often described in one direction: a person gives an agent a goal. In practical systems the direction reverses repeatedly. An agent asks a person to provide missing information, approve an irreversible action, resolve a conflict between policies, supply credentials, perform a physical manipulation, or accept responsibility for a judgment the system cannot make. This is better understood as \emph{agent-to-human escalation} than as full delegation of a job to a human. Among the artifacts examined through the cutoff date, none specified a general protocol for recruiting, compensating, prioritizing, and verifying arbitrary human work across institutions.

Pieces of the interaction exist. MCP elicitation introduced server-initiated requests for structured user input, with the client controlling how the request is presented and allowing acceptance, rejection, or cancellation \citep{mcp2025elicitation}. The later multi-round-trip mechanism generalized mid-call requests in a stateless transport \citep{mcp2026release}. A2A's \texttt{input-required} state supports additional input during a persistent task \citep{a2a2026spec}. Magentic-UI provides a richer research prototype: co-planning, co-tasking, multi-task management, action guards, memory, and final-answer verification around a multi-agent web and coding system \citep{mozannar2025magenticui}. Its technical report combines benchmark evaluation, simulated users, qualitative studies, and targeted safety tests; these are informative but do not show that approval-heavy supervision remains effective over long periods or under high task volume.

The reverse direction also began to move beyond clarification interfaces. RentAHuman's first-party documentation exposes an MCP server and REST API through which an authenticated agent can search for people, create a bounty, accept an applicant, inspect submitted evidence, and release escrowed payment \citep{rentahuman2026docs}. This is an emerging marketplace interface, not a general protocol or evidence that agent-directed labour is broadly safe. A February 2026 preprint measured 303 bounties on the same marketplace and attributed 99 (32.7\%) to API-key or MCP channels. Its dual-coder analysis identified credential fraud, identity impersonation, reconnaissance, social-media manipulation, authentication circumvention, and referral fraud among posted tasks \citep{mehta2026hiring}. The study is a single-platform, single-snapshot preprint and its retrospective screening experiment does not establish causal effects or ecosystem-wide prevalence. It nevertheless demonstrates why agent-to-human execution is more than a user-interface feature: an agent can purchase physical presence, identity-dependent access, or social influence that bypasses the technical boundary of its own runtime.

Human involvement should consequently be evaluated as a control channel with its own failure modes. Approval prompts can be too frequent, too technical, or timed after the meaningful decision has already been made. People can habituate, misunderstand the represented risk, or rubber-stamp a recommendation that the agent framed confidently. Conversely, requiring confirmation for every action can eliminate the value of delegation. Market-mediated delegation adds worker consent, compensation, identity, jurisdiction, prohibited-task screening, evidence integrity, dispute resolution, and spending authority. The research problem is to elicit human judgment where it changes the decision boundary, automate checks that are faster and more reliable in code, and prevent paid human action from becoming an unmonitored capability-escalation path.

\subsection{Multi-agent systems trade modularity for coordination risk}

Multi-agent architectures divide work among roles such as planner, researcher, coder, critic, or executor. AutoGen made such conversational organizations programmable and supported agents backed by models, tools, or people \citep{wu2024autogen}. Magentic-One later coupled an orchestrator to specialized web, file, coding, and terminal agents in an open technical-report system \citep{fourney2024magenticone}. This organization can isolate tools, reduce context competition, enable parallelism, and make responsibility boundaries explicit. It can also spend more inference, repeat evidence, propagate one model's error through several roles, and obscure which component made a consequential choice.

The empirical question is not whether multiple named roles sound appropriate, but whether they improve a matched-budget baseline on the relevant task. A 2026 preprint found that single-agent systems matched or outperformed several multi-agent organizations on multi-hop reasoning when reasoning-token budgets were held constant, while identifying API and benchmark artifacts that can inflate apparent gains \citep{tran2026singleagent}. The result is limited to the tested models and reasoning tasks, yet it establishes a necessary evaluation principle: compare equal computation and report coordination overhead. Peer-reviewed evidence on multi-agent debate adds a different warning. Becker and colleagues found that discussions can drift from the original problem over turns; human analysis attributed common cases to lack of progress, low-quality feedback, and lack of clarity \citep{becker2026drift}. More dialogue is not monotonically better reasoning.

Multi-agent systems are most defensible when decomposition corresponds to genuine heterogeneity: distinct permissions, models, sensors, institutional owners, or independently verifiable expertise. Multiple copies of the same model prompted with different role names may increase search diversity, but they do not automatically provide independent evidence. Shared training data, prompts, tools, and retrieval sources create correlated error. A critic that sees the same misleading context as the executor is not an external verifier.

\subsection{Persistence is not one capability}

Persistent agency is likewise a bundle. \emph{Interaction persistence} retains conversational state over turns. \emph{Task persistence} lets work pause, resume, or continue asynchronously. \emph{Episodic memory} stores observations and reflections across encounters. \emph{Skill persistence} preserves executable procedures learned from prior tasks. \emph{Environmental persistence} means that world state continues to evolve or remains changed when outside the agent's current view. \emph{Authority persistence} retains credentials or permission to act after the original interaction.

Generative Agents illustrates episodic and environmental persistence in a controlled social simulation. Twenty-five language-model agents maintained memory streams, generated reflections, planned daily activities, and interacted in a Sims-like environment; ablations connected memory, reflection, and planning to judged believability \citep{park2023generativeagents}. Voyager illustrates skill persistence: an automatic curriculum, an executable code library, environmental feedback, and iterative repair supported open-ended exploration in Minecraft \citep{wang2023voyager}. Coding products and A2A tasks illustrate resumable task persistence \citep{openai2025codexga,a2a2026spec}. None of these, by itself, grants legitimate recurring authority.

This distinction is crucial because durable permissions create a qualitatively different risk from durable memory. A scheduled agent that can repeatedly modify live systems may encounter changed policies, expired assumptions, or new affected parties. It needs explicit renewal conditions, revocation, budget limits, checkpoints, and audit records. Long-running-agent benchmarks are beginning to target monitoring and asynchronous work, but the public evidence examined here is substantially stronger for completing bounded episodes than for managing obligations over weeks or months. Persistence should therefore be reported object by object, not inferred from the phrase ``long-term agent.''

\section{Persistent virtual environments and world models}

\subsection{Agents and world models occupy different sides of the loop}

The phrase ``world model'' is used for systems with materially different representations: a predictive latent model used for planning, an action-conditioned video generator, an explicit structured simulation, or a platform that produces interactive environments. These systems sit primarily on the environment or prediction side of Figure~\ref{fig:system}. An agent, by contrast, maintains or receives an objective and selects actions. A world model may help an agent anticipate consequences or supply training experience, but it is not evidence of goal-directed autonomy unless an evaluated agent actually closes the loop through it.

This distinction clarifies several influential results. The SIMA project trains an instructable agent across research environments and commercial three-dimensional games using screen images and language as inputs and keyboard--mouse actions as outputs. The 2024 technical report emphasizes a generic human-like interface and cross-environment generality, while its initial task set concentrates on basic skills typically completed over short time scales \citep{sima2024}. Voyager instead uses an automatic curriculum, environment feedback, iterative code generation, and a growing executable skill library to explore Minecraft \citep{wang2023voyager}. Both are agents in simulated environments. Their results support language grounding, transfer, exploration, and persistence claims within the evaluated worlds; they do not establish physical robustness or authority in live social systems.

World-foundation-model projects target the complementary substrate. NVIDIA's Cosmos preprint describes video curation, tokenization, pretrained world foundation models, and post-training components intended for physical-AI development \citep{nvidia2025cosmos}. V-JEPA~2 learns predictive representations from more than one million hours of video and then post-trains an action-conditioned model with fewer than 62 hours of unlabelled robot video; the report demonstrates zero-shot image-goal planning on Franka arms in two laboratories \citep{assran2025vjepa2}. These results connect representation learning to planning more directly than visually plausible generation alone. They remain preprint evidence, and the demonstrated manipulation tasks do not test long-lived objectives, heterogeneous hardware, or open human environments.

\subsection{Visual continuity is not persistent world state}

Generative interactive worlds expand the range of environments in which agents could learn or be evaluated. Google DeepMind's Genie~3 preview reports text-conditioned environments navigable at 24 frames per second and 720p, with largely consistent interaction over several minutes and visual memory extending to approximately one minute \citep{deepmind2025genie3}. The same announcement explicitly limits direct agent actions, multi-agent interaction, geographic accuracy, text rendering, and interaction duration, and classifies access as a limited research preview. It also states that Genie simulates consequences from actions without knowing the attached agent's goal. This is exactly the environment--agent separation required for careful interpretation.

Pure pixel recurrence has a specific limitation: what is not visible may be compressed into a finite recent context rather than stored as an independently queryable entity. A scene can look coherent over a camera trajectory while failing object permanence after long occlusion, allowing incompatible views, or changing causal variables without a stable record. Evaluation should therefore distinguish at least four properties: short-range visual continuity, camera-independent spatial consistency, persistent object identity and attributes, and rule-consistent state transitions under intervention.

VAST AI Research's Project Eden is a useful frontier example because its stated design targets this distinction. The June 2026 first-party research preview describes a persistent multiplayer world model that separates an evolving structured, camera-independent world state from a state-to-observation interface and generative neural rendering \citep{vast2026eden}. The stated architecture is relevant to this review's distinction: off-screen objects and changes could, in principle, remain in shared state rather than being reconstructed only from recent pixels, and multiple users or agents could act on that state. The source also identifies richer physics, larger environments, broader free-viewpoint exploration, finer object interaction, stronger transition models, and more extensive evaluation as future work.

Project Eden should therefore appear in the review as an evidence point at the frontier of environmental persistence, not as its main thesis. At the cutoff it was a company research preview, not a peer-reviewed paper, an open model, or an independently reproduced agent benchmark. Its ``first'' wording is a first-party positioning claim that this review does not adopt. The evidence supports the existence and stated architecture of a preview; it does not yet support general claims about long-horizon consistency, multi-agent learning benefit, or autonomous agents.

\subsection{Generated 3D artifacts and interactive worlds require separate evaluation}

Another line of work asks agents to construct three-dimensional environments rather than merely act within them. VibeWorlding, a 2026 preprint, evaluates multimodal agents that create 3D open-world artifacts end to end \citep{ning2026vibeworlding}. This task combines coding, asset creation, spatial composition, and iterative visual feedback. It is evidence about agentic artifact production in a sandbox. Calling the resulting artifact an ``open world'' does not make the execution environment open in the governance sense: the build process can still be bounded, reset, and inspected.

This separation suggests a more informative evaluation matrix for world-acting research. Environment models should be tested for counterfactual accuracy, state persistence, causal consistency, controllability, and coverage of failure-relevant events. Agents should be tested for goal completion, exploration efficiency, recovery, policy compliance, and generalization. The joint system should be tested for model exploitation: an agent may learn shortcuts that succeed in the learned world but fail under the target environment's dynamics. A generated world that is diverse but causally wrong can make training less rather than more useful.

\subsection{Why virtual environments remain indispensable}

These limitations do not diminish the scientific value of virtual environments. Simulators make dangerous or rare events repeatable, permit controlled counterfactuals, expose hidden state to evaluators, and support large-scale curriculum generation. Persistent shared worlds also provide a tractable setting for studying coordination, memory, norms, and resource competition without imposing the corresponding external costs. The error is not using a simulator; it is forgetting which uncertainties the simulator removed.

A credible path toward physical or social deployment therefore requires staged transfer. First establish competence and recovery in a fully inspectable environment. Then introduce observation noise, delayed effects, other agents, non-stationary rules, and partial resetability. Validate a limited real interface behind hard constraints before expanding action scope. World models can accelerate each stage by generating scenarios or predicting outcomes, but the final evidence must come from the environment whose risks the deployment will bear.

\section{Physical interfaces, robotics, and autonomous laboratories}

\subsection{Physical action changes the validation problem}

When an agent acts through a physical device, mistakes are no longer confined to an information state. Perception is partial and noisy; calibration drifts; actuators have latency and tolerance; materials deform, spill, wear, or react; and people may share the workspace. The same high-level instruction can require different safe trajectories on nominally identical equipment. Reverting a software snapshot has no general physical analogue. Physical agency therefore demands evidence about both the selected objective and the execution controller that realizes it.

Vision-language-action (VLA) models connect semantic instructions and visual observations to robot actions. RT-2 co-fine-tuned vision-language models on web-scale vision-language tasks and robot trajectories, representing actions as tokens. Its peer-reviewed study reported 6,000 evaluation trials and improved generalization to novel objects and instructions, including rudimentary semantic reasoning \citep{zitkovich2023rt2}. OpenVLA trained a 7-billion-parameter open VLA model on 970,000 robot episodes and evaluated it across multiple robot embodiments and manipulation suites \citep{kim2024openvla}. The Gemini Robotics technical report described a family of VLA and embodied-reasoning models evaluated on varied manipulation, novel objects and environments, and adaptation to additional robot embodiments \citep{gemini2025robotics}.

These systems are important evidence that foundation-model representations can improve the generality of robot policies. They should not automatically be classified as persistent autonomous agents. A VLA may map a current instruction and observation to a short sequence of motor actions without owning the high-level objective, retaining authority, or deciding when an experiment should be stopped. The orchestration layer that selects tasks, interprets safety constraints, and verifies outcomes may remain separate. Reporting should identify which layer was evaluated and whether the action distribution included people, fragile objects, unexpected obstacles, or faults.

\subsection{A hardware standard is an interface claim}

The Model Hardware Standard (MHS) makes this layered architecture explicit. Anthropic's 27 August 2026 announcement describes a limited research preview, developed initially with HHMI Janelia, that proposes a model-agnostic driver for programmable laboratory and manufacturing equipment \citep{anthropic2026mhs}. Devices expose discoverable state and simple read/write-style primitives; natural-language tags describe characteristics and enforced safety limits; and agents can access drivers through MCP, command-line tools, or code. The announcement presents proof-of-concept work spanning microscopes, liquid handlers, robotic arms, qPCR monitoring, assay optimization, and laser alignment.

MHS is relevant here as an interface proposal because it joins the digital tool layer to physical instrumentation through a common abstraction. It is not evidence that physical autonomy has been standardized. At the cutoff, access was limited, the announcement preceded a promised open-source release, and the reported demonstrations were supplied by Anthropic and participating laboratories rather than an independent conformance program. The source itself describes further work on safety evaluations and best practices, and partner accounts identify failures that still required expert physical insight. A common driver can reduce integration effort and expose safety metadata; it cannot infer accurate limits, guarantee that tags are complete, or make a model understand fluid dynamics and collision risk.

The correct comparison is therefore between MHS and earlier software action interfaces, not between MHS and a general-purpose robot brain. Like MCP, it defines how capabilities can be exposed. Unlike a purely digital tool protocol, its commands ultimately cross a physical boundary. This raises additional requirements: device-certified interlocks, units and calibration metadata, freshness and uncertainty for sensor values, idempotency declarations, simulation or dry-run modes, emergency stops independent of the model, and append-only records linking high-level intent to low-level commands.

\subsection{Autonomous laboratories supply bounded closed-loop evidence}

Scientific laboratories provide some of the clearest demonstrations of language-model agents affecting the physical world because experiments already have instruments, protocols, measurements, and expert comparison. Coscientist combines GPT-4 with literature and documentation search, code execution, and experimental automation. Its Nature paper evaluated six tasks, including optimization of palladium-catalysed cross-coupling reactions, and characterized the system as supporting semi-autonomous experimental design and execution \citep{boiko2023coscientist}. ChemCrow couples a language model to 18 chemistry-oriented tools and demonstrated planning and execution of selected syntheses under an expert-designed workflow \citep{bran2024chemcrow}. These studies establish that tool-augmented language systems can close parts of a design--make--test--analyze loop in a bounded apparatus.

They do not establish an unattended general scientist. The research goal and available tools are prescribed; chemical access and execution are constrained; a small number of demonstrations cannot estimate rare hazards; and expert judgment remains essential for interpreting whether a procedure is scientifically meaningful. Scientific correctness also has several levels. A system may execute the requested protocol, produce a valid measurement, draw an internally consistent conclusion, and still ask the wrong research question or overlook an alternative explanation. Instrument success and scientific validity require different evaluators.

The Artificially Intelligent Lab Assistant (AILA) and AFMBench extend this evaluation emphasis to atomic-force microscopy. The 2025 Nature Communications study evaluated single- and multi-agent workflows across experimental design, operation, and analysis, finding that domain question-answering performance did not reliably translate into laboratory competence. It also reported sensitivity to prompt formatting and instruction deviations termed ``sleepwalking,'' despite multi-agent gains in the evaluated setting \citep{mandal2025afm}. This is unusually direct evidence for the review's central claim: access to a complete experimental workflow can advance faster than robust alignment to the workflow's instructions.

Autonomous laboratories also predate the current language-agent framing. A-Lab integrated robotics, databases, machine-learning interpretation, text-mined synthesis knowledge, and active learning for inorganic-material synthesis \citep{szymanski2023alab}. A January 2026 author correction revised claims about material novelty and synthesis outcomes; we therefore use A-Lab here as evidence of system integration, not as an unqualified count of novel materials \citep{szymanski2026correction}. This lineage shows that closed-loop autonomy depends on far more than an LLM. Reliable schedulers, calibrated instruments, machine-readable samples, experimental-design algorithms, and domain-specific error handling supply much of the actual autonomy. Language models add flexible interfaces, knowledge access, and plan composition, but inherit rather than replace those engineering requirements.

The ADePT perspective offers a useful complementary vocabulary for laboratory robotics: adaptability and learning, dexterity, perception, and task complexity \citep{salazarvillacis2026adept}. Its analysis treats interoperability as orthogonal to capability, which fits the distinction made here between MHS-style interfaces and robot proficiency. A laboratory may be highly integrated but mechanically narrow, or dexterous but difficult to connect to an agent. Evaluation should report both.

\subsection{From demonstration to defensible physical deployment}

Physical evidence should progress through a staged assurance case. First, test component policies against recorded or simulated state, including adversarial and out-of-distribution inputs. Second, run hardware-in-the-loop trials with inert materials, conservative limits, and independent state measurement. Third, evaluate bounded procedures under expert supervision with preregistered stop conditions and sufficient repetitions to characterize ordinary failures. Fourth, introduce fault injection, sensor disagreement, degraded calibration, and recovery. Only then should unattended duration, device diversity, or experimental freedom expand.

No single aggregate ``autonomy level'' captures these requirements. A system may plan novel experiments but require manual transfer; another may run an instrument around the clock but execute a fixed protocol; a third may coordinate devices while asking a person to approve every scientific decision. Reports should separately state who chooses the goal, who converts it to actions, what the physical controller can do, which constraints are enforced outside the model, how success is measured, and who bears responsibility for unexpected consequences.

The physical frontier thus strengthens rather than overturns the review's thesis. Robotics, laboratories, and MHS show that language-conditioned systems are acquiring broader action channels. Their most credible results arise where the environment is bounded, instrumentation is structured, and external checks are strong. As those boundaries relax, the burden of proof should rise faster than the action space.

\section{Reliability, security, and governance across environments}

\subsection{Outcome verification is the missing middle}

The evaluations examined here variously measure two endpoints: whether the model selected a plausible action, and whether the final task state matched a benchmark predicate. Safe delegation requires a chain between them. The request must be interpreted correctly; the principal must possess authority to request it; each action must be permitted under current conditions; the environment must execute it as assumed; the final state and important side effects must be observed; and failures must trigger containment or recovery. An agent's fluent explanation does not verify any link in this chain.

Deterministic checks are strongest where state is structured. Database fields, repository diffs, test results, checksums, device interlocks, and transaction receipts can provide evidence independent of a model's narrative. Semantic goals---``make this document accurate,'' ``treat this person fairly,'' or ``choose a scientifically informative experiment''---cannot generally be reduced to one predicate. They need layered evaluation: machine-checkable invariants, domain-specific tests, independent models or tools where appropriate, and accountable human judgment for residual ambiguity. Using the acting model as the only judge creates correlated error and an incentive to rationalize its own trajectory.

The verifier is also an attack surface. An agent can optimize a proxy, change the object being evaluated, or select evidence that makes an incorrect action look successful. In a learned virtual world, it may exploit model error; in a web task, it may satisfy the visible page while modifying the wrong account; in a laboratory, a sensor can report a plausible number after a calibration fault. Verification must therefore bind the objective, action log, environment identity, and measurement provenance before execution wherever possible.

\subsection{Prompt injection becomes an action-security problem}

Retrieval and computer use collapse the boundary between data and instructions. A web page, document, tool output, or message from another agent can contain text that attempts to redirect the model. Once the model has credentials or an action interface, indirect prompt injection is no longer only a content-integrity problem; it can cause data disclosure or unauthorized state change. AgentDojo provides a dynamic environment for evaluating such attacks and defenses under utility constraints, while ToolEmu uses an LM-emulated sandbox to identify risky behaviours across tool-use scenarios \citep{debenedetti2024agentdojo,ruan2023toolemu}. Agent Security Bench broadens formalized attack and defense evaluation across agent components \citep{zhang2024asb}.

These benchmarks are valuable because they evaluate a system under adversarial context rather than ask a base model a safety question. Their coverage is necessarily partial. Attack strings, tool schemas, model endpoints, and defense prompts change, and an LM-emulated environment cannot reproduce every external side effect. A defense that lowers attack success by refusing many benign actions may also destroy utility. Security claims should consequently report both task performance and attack outcomes, along with model and harness versions, permissions, inference budget, and the exact trust boundary.

An Anthropic incident report published on the cutoff date described pre-release models, deliberately evaluated without cyber safeguards, taking unauthorized actions on the live internet when a third-party environment was misconfigured or intentionally internet-enabled. This first-party report establishes boundary failures under those particular tests, not their frequency in ordinary deployments \citep{anthropic2026alignmentsecurity}.

\subsection{Authority must be represented outside natural language}

Natural-language instructions are expressive but underspecified as access-control policies. ``Handle my travel'' does not determine a spending ceiling, acceptable airports, whether a non-refundable purchase is allowed, or who may see identity documents. ``Run the experiment'' does not define chemical quantities, maintenance windows, or which anomaly requires evacuation. An authorization layer should translate delegated intent into machine-enforceable capabilities: named resources, permitted operations, value and time limits, conditions for renewal, and requirements for approval or dual control. Work on authenticated and auditable delegation formalizes this need, although proposed frameworks remained early research rather than widely validated infrastructure at the cutoff \citep{south2025delegation}.

MCP and A2A supply useful security hooks but do not remove this application responsibility. OAuth scopes, authenticated Agent Cards, signed metadata, and per-service policy can identify and constrain parties \citep{mcp2026spec,a2a2026spec}. The harness must still decide which model process receives which credential, prevent confused-deputy behavior, and revoke access when a task ends. Agent-to-human marketplaces additionally require spending limits and prohibited-task policies; MHS-controlled devices require independent hardware limits. The more general the interface, the less safe it is to infer authority from the fact that an operation is technically available.

\begin{table*}[t]
\centering
\caption{Assurance obligations grow with persistence and environmental coupling. The examples are controls to evaluate, not claims that current systems implement them completely.}
\label{tab:assurance}
\scriptsize
\begin{tabularx}{\textwidth}{L{2.3cm}Y Y Y Y}
\toprule
\textbf{Environment} & \textbf{Typical evidence of success} & \textbf{Characteristic hidden state} & \textbf{Required control emphasis} & \textbf{Residual question} \\
\midrule
Tool/API sandbox & Returned object, state predicate, replayable trace & Service policy and effects outside the schema & Typed arguments, idempotency, allowlists, deterministic evaluators & Did the valid call satisfy the user's actual intent? \\
Live web/desktop & Account state, receipt, artifact, independent query & Focus, overlays, remote policy, other users, indirect prompts & Isolated browser, credential scoping, confirmations, transaction reconciliation & What affected-party or account consequence was omitted? \\
Agent or human delegation & Signed artifact, task record, evidence package, acceptance decision & Contractor competence, incentives, private actions, copied context & Identity, bounded brief, budget, provenance, conflict and dispute mechanisms & Is ``completion'' truthful, lawful, and independently checkable? \\
Generated/\allowbreak simulated world & Instrumented state, causal probes, repeatable episode & Model shortcuts and mismatch with the target world & State access for evaluators, counterfactual tests, held-out dynamics, transfer validation & Did the agent learn the task or exploit the simulator? \\
Robot or laboratory & Independent sensors, calibrated measurement, physical inspection & Wear, contamination, occlusion, people, irreversible change & Interlocks, units, uncertainty, staged trials, emergency stop, expert accountability & Is failure bounded when the model or instrument is wrong? \\
\bottomrule
\end{tabularx}
\end{table*}

\subsection{Human oversight must be designed and measured}

``Human in the loop'' is not a binary safeguard. The person may set a goal, approve a plan, confirm a transaction, monitor a stream, intervene during execution, inspect the final artifact, or investigate an incident. Each position sees different information and imposes different latency and workload. A user asked to approve a high-level plan cannot catch a low-level argument error; an operator shown every tool call may miss the one consequential call among hundreds. Oversight evaluation should measure detection, intervention quality, time, cognitive load, and recovery outcome---not merely the presence of an approval widget.

Magentic-UI's co-planning, co-tasking, action guards, and final verification constitute a useful research platform for these questions \citep{mozannar2025magenticui}. MHS proof-of-concept descriptions similarly include cases in which an agent asks a researcher whether to stop a procedure and cases in which expert guidance was needed to diagnose a physical failure \citep{anthropic2026mhs}. These examples suggest that high-value involvement often occurs at decision boundaries and anomalies rather than every routine action. They do not establish the correct threshold, especially when operators supervise many concurrent agents.

Oversight also requires meaningful alternatives. A person must be able to inspect evidence, reject or modify a plan, pause execution, revoke credentials, and restore a safe state. If an opaque agent has already committed an irreversible action before asking, the interaction is notification rather than control. For affected people who are not the delegator, governance further requires notice, contestability, and an accountable institution; a technically successful task can still allocate burdens unfairly or violate rights.

\subsection{A minimum reporting standard for world-acting systems}

Comparability would improve if every agent evaluation reported six objects. First, identify the exact model endpoint or weights, decoding and reasoning budget. Second, describe the harness: prompts, context policy, memory, tools, retries, subagents, and verifiers. Third, state the environment's reset conditions, network access, credentials, and whether other people or live resources were affected. Fourth, separate attempted, technically completed, policy-compliant, and externally verified outcomes. Fifth, report repeated trials, costs, latency, intervention frequency, and failure severity rather than only best-case success. Sixth, archive action traces and artifacts subject to privacy and security constraints.

For multi-agent systems, equalize total inference and tool budgets against a strong single-agent baseline. For persistent systems, report duration, interruptions, state migration, credential renewal, and memory deletion. For world models, report causal and off-screen state consistency rather than curated videos alone. For physical systems, provide device configuration, calibration, independent sensing, interlocks, incident definitions, and sufficient repetitions. Company previews can contribute valuable architecture and limitation disclosures, but should remain visually and verbally separate from independently peer-reviewed evidence.

These requirements shift evaluation from ``can the agent act?'' to ``under what configuration, authority, evidence, and failure boundary should the action be delegated?'' That is the minimum unit of knowledge needed for cumulative progress.

\section{A research agenda for justified delegation}

A useful research program does not require a single autonomy score. It requires evidence that action scope can be increased without losing control of intent, authority, state, and recovery. We use \emph{justified delegation} as an analytical and normative heuristic for the largest class of tasks for which a configured system can demonstrate adequate competence, respect a bounded authorization, expose evidence of outcomes and side effects, and return to a safe state when assumptions fail. It is not an empirically measured law, benchmark, or certification; it is a decision criterion for asking when broader authority is defensible. Seven research priorities follow from the preceding synthesis.

\subsection{Evaluate the model and harness as a coupled intervention}

Model releases are easier to name than harness changes, so benchmark narratives often attribute system gains to the model. Future comparisons should use factorial designs where feasible: vary the model, interface, memory policy, verifier, and retry budget independently. At minimum, hold the harness fixed when comparing models and hold the model fixed when claiming a harness contribution. Report total inference and tool cost, including critic and subagent calls. This is especially important when comparing single- and multi-agent systems, because additional roles can silently purchase more search.

Harness research should prioritize interpretable state over ever-longer transcripts. Plans, commitments, uncertainties, tool results, and authorization decisions should be represented as typed objects with provenance and explicit invalidation conditions. Context compression should preserve why a fact is believed and when it expires, not merely a fluent summary. A verifier should receive the evidence needed to challenge the actor, not inherit all of its assumptions by default.

\subsection{Make authority capability-based, renewable, and revocable}

Agent protocols should carry task-scoped capabilities rather than ambient credentials. A capability should name the resource, allowed operation, parameter or value bounds, expiration, delegability, and required approval. Subdelegation through A2A or a human marketplace should attenuate rather than amplify authority: a downstream agent or worker should receive no more permission than the principal intentionally granted. Every handoff should preserve the chain from human or institutional principal to final executor.

Natural-language intent will remain necessary, but it should compile into a reviewable policy object. Research is needed on detecting ambiguity during compilation, reconciling policies from several principals, and representing non-delegable decisions. Protocol conformance tests should include adversarial authorization cases, confused-deputy scenarios, replayed task handles, and revocation during a long-running action. Adoption metrics should distinguish implementations that expose a protocol endpoint from those that pass such tests.

\subsection{Treat human participation as a scarce, safety-critical resource}

Human--agent research should optimize the allocation of judgment, not minimize the count of human actions. Systems need calibrated triggers for asking, approving, observing, intervening, and auditing. Experiments should measure whether a person detects a planted error, whether the information display supports a correct decision, how supervision quality changes with concurrency and fatigue, and whether users can reconstruct responsibility after an incident.

Agent-to-human marketplaces require an additional research program. Task filters must reason about combinations of benign-looking steps, and worker-provided evidence must be authenticated without creating invasive surveillance. Economic and social questions include wage setting, informed consent to agent clients, discriminatory task allocation, reputation manipulation, jurisdiction, and who is liable when the agent's brief is unlawful or dangerous. The early RentAHuman measurement shows both programmatic use and active abuse categories, but replication across platforms and time is required before prevalence can be generalized \citep{mehta2026hiring}.

\subsection{Benchmark persistence as controlled state change over time}

Long-task evaluation should go beyond increasing the maximum number of steps. A persistent benchmark should include interruptions, stale observations, policy changes, expiring credentials, concurrent actors, delayed outcomes, and tasks that should be cancelled rather than completed. It should score state integrity, resource consumption, forgotten obligations, safe resumption, and appropriate reauthorization. The unit of success should include a durable evidence record, not only an endpoint reward.

Memory needs separate tests for recall, provenance, correction, and deletion. A system that remembers a false inference consistently is persistent but unsafe. A system that cannot forget a revoked preference or sensitive record also fails. Skill libraries such as Voyager's motivate tests of whether a stored procedure remains valid after the environment changes and whether the agent knows when not to reuse it \citep{wang2023voyager}.

\subsection{Connect world models to causal and transfer evidence}

World-model evaluation should move from visual demonstrations to intervention-sensitive measurement. Benchmarks should hide objects for long intervals, revisit locations from conflicting viewpoints, alter one causal variable while holding appearance constant, and test conservation or task-specific rules. Shared-world benchmarks should introduce independent agents with incompatible information and measure whether the state remains coherent under concurrent actions. Project Eden's explicit-state proposal makes these questions concrete, but open artifacts and independent tests are needed \citep{vast2026eden}.

For agent training, diversity is useful only when paired with fidelity relevant to the downstream task. Researchers should report which behaviours transfer from a generated environment, which exploit generator artifacts, and how uncertainty in the model is propagated to the planning policy. A staged ``simulation--hardware-in-the-loop--bounded deployment'' protocol can localize failures before physical exposure. Learned world models should abstain or defer to direct sensing when predictive uncertainty crosses a task-specific threshold.

\subsection{Build physical assurance cases from independent layers}

Physical-agent research should combine, but not collapse, three evaluations: semantic planning, device-level execution, and system-level safety. A high-level agent may propose a valid plan while the controller cannot execute it; a robot may execute accurately while pursuing an invalid goal. Independent interlocks and emergency stops should remain effective even if the model, network, or orchestration process fails. Sensor disagreement, calibration drift, wear, occlusion, and human entry into the workspace should be routine test conditions rather than afterthoughts.

MHS creates an opportunity for shared device manifests and conformance suites if its promised open-source development yields stable public artifacts \citep{anthropic2026mhs}. Useful extensions would include standardized units, uncertainty and freshness metadata, command preconditions and reversibility, dry-run semantics, hazard classifications, and signed calibration records. The standard's value should be evaluated by integration time, cross-device portability, error detection, and safe recovery across independent laboratories---not only by whether a model can issue a common command.

\subsection{Replace benchmark snapshots with longitudinal evidence}

Finally, world-acting systems need post-deployment evidence. Benchmark results should be linked to model and harness versions so that regressions are visible. Deployments should collect privacy-preserving denominators: how many tasks were attempted, declined, escalated, corrected, reversed, or abandoned; how severe failures were; and how much human work was required. Incident taxonomies should connect prompt injection, permission misuse, coordination failure, memory error, interface drift, and physical fault to the layer that could have prevented propagation.

The appropriate endpoint is not zero human involvement or indefinite runtime. It is a system whose boundaries remain understandable as capability changes. A narrow agent with strong evidence may deserve broad operational use within its scope; a more capable agent may deserve less authority when its actions are hard to verify. This reverses the usual presumption that higher benchmark performance should automatically unlock a larger action surface.

\subsection{Limits of the present synthesis}

This review is intentionally selective and cannot support bibliometric prevalence claims. The cutoff captures an unusually fast-moving period; protocol revisions, product availability, and benchmark rankings can change after 31 August 2026. Several frontier claims---notably Genie~3, Project Eden, and MHS---derive from first-party previews without archival papers or independent replications at the cutoff. We retain them because they expose important architectural directions, but separate their evidential status throughout.

Coverage is also asymmetric. The public evidence base examined here is concentrated in software engineering, web tasks, games, robotics laboratories, and English-language interfaces, settings with accessible benchmarks and artifacts. Evidence from high-stakes institutions, long-running enterprise operations, and affected non-users is scarcer within this review or not publicly auditable. We therefore do not infer that an absent public failure mode is rare. Nor do we treat the taxonomy as a developmental sequence: digital, social, virtual, and physical coupling overlap, and a system may be mature on one assurance dimension while immature on another.

\section{Conclusion}

Agentic AI is best understood as a change in control architecture, not a label attached to a language model. Models now participate in systems that call tools, operate computers, delegate to agents and people, retain work over time, interact with generated worlds, and control physical devices. Across the studies and artifacts examined here, the strongest evidence supports expanded action coverage and meaningful but uneven gains in task competence. The same body of public evidence does not establish a general transition to reliable, unattended open-world autonomy.

Separating model, harness, environment, and delegator resolves much of the apparent contradiction. MCP and A2A standardize interfaces but not trustworthy decisions. Multi-agent organization can provide real modularity but not automatic independence or efficiency. World models can supply persistent and predictive environments but are not themselves agents. VLA and laboratory studies provide bounded evidence of increasingly direct physical coupling, whereas the MHS preview proposes a more general driver layer; both make external constraints, calibration, and recovery more important.

We therefore use justified delegation as a practical analytic heuristic: ask how much state-changing authority can be granted with evidence that the system will preserve intent, remain within permission, reveal what it changed, and fail safely. Advancing that frontier requires coupled model--harness evaluation, typed and revocable authority, measured human control, causal tests of persistent environments, layered physical assurance, and longitudinal deployment records. The aim is not maximal independence or compliance with a newly invented score. It is accountable action whose scope grows only when the relevant evidence makes delegation defensible.

\section*{Use of generative AI}

Generative artificial intelligence assisted literature discovery, drafting, language editing, translation, and \LaTeX{} preparation; the authors verified the sources and remain responsible for the manuscript.

\bibliographystyle{plainnat}
\bibliography{references}

\end{document}